%% file: main.tex
\documentclass[sigconf]{acmart}
\renewcommand\footnotetextcopyrightpermission[1]{} 
\AtBeginDocument{%
  }

\usepackage{algorithm}
\usepackage{listings}
\usepackage{amsmath} 
\usepackage{algpseudocode}
 \usepackage{xcolor}
 \usepackage{hyperref}
 \usepackage{xurl}
 \usepackage{enumitem}
 \usepackage{comment}
\begin{document}

\newcommand{\FlashV}{FlashAttention-V}

\title{FlashAttention for Scalable Vector Architectures}


\author{Sonia Rani Gupta}
\affiliation{%
  \institution{Chalmers University of Technology}
  \city{Gothenburg}
  \country{Sweden}}
\email{soniar@chalmers.se}

\author{Nikela Papadopoulou}
\affiliation{%
  \institution{University of Glasgow}
  \city{Glasgow}
  \country{UK}}
\email{nikela.papadopoulou@glasgow.ac.uk}

\author{Miquel Peric\`as}
\affiliation{%
  \institution{Chalmers University of Technology}
  \city{Gothenburg}
  \country{Sweden}}
\email{miquelp@chalmers.se}
\begin{abstract}
 \input{abstract}

\end{abstract}

\pagestyle{empty}

\maketitle

\section{Introduction}
Transformer-based models underpin modern AI systems, powering applications ranging from chatbots and coding assistants to machine translation and AI agents, spanning a wide range of scales. 
Large Language Models (LLMs), with hundreds of billions of parameters, are typically deployed in data centers, whereas Small Language Models (SLMs), in the 1-10B parameter range, are optimized for edge deployment~\cite{hasan2026assessingsmalllanguagemodels, OrpekLLMSLM2024, lu2024SLMs}.
Across this spectrum, transformer inference demands high throughput, low latency, and strong energy efficiency, from large-scale server deployments to battery-constrained edge platforms~\cite{AminabadiDeepSpeedinferenceSC22, energyefficiencyInference}.

The attention module is a core component of all transformer architectures~\cite{vaswani2017attention}.  
 Multi-head attention (MHA) extends the original formulation by computing multiple attention heads in parallel, enabling the model to capture diverse relational patterns. 
To reduce the memory and bandwidth overhead of storing the key-value (KV) cache during inference, variants such as Multi-Query Attention (MQA) and Grouped-Query Attention (GQA) share keys and values across query heads, reducing the KV cache footprint~\cite{ainslie-etal-2023-gqa, shazeer2019fasttransformerdecodingwritehead, lu2024SLMs}.

Efficient execution of these attention variants has been explored across diverse hardware platforms. Attention implementations have been optimized for GPUs~\cite{pytorch2_2_blog, llama_cpp, huggingface_transformers, vllm} and dedicated AI accelerators~\cite{jax_pallas, splashattention, vllm} to maximize throughput for both training and inference. They have also been explored on CPUs to improve transformers inference due to the CPUs' high availability, low latency, and portability across edge devices and data centers~\cite{onednn_sdpa, llama_cpp,vllm}. 

Although these variants differ in how queries, keys, and values are organized, they all rely on the same underlying self-attention computation.
Self-attention requires computing and storing large intermediate matrices, such as score matrix $S=QK^T$, where $Q$ and $K$ represent Queries and Keys, as well as the resulting softmax attention weights. This leads to significant memory capacity and bandwidth overhead, which becomes particularly pronounced as the sequence length grows.
To address this memory bottleneck, Rabe and Staats~\cite{rabe2022selfattentiondoesneedon2} introduced a memory-efficient attention algorithm with online softmax, enabling exact self-attention without storing the full intermediate score matrix S. Building on this foundation, FlashAttention~\cite{dao2022flashattentionfastmemoryefficientexact, dao2023flashattention2fasterattentionbetter} further optimizes execution by using tiling and fused kernels to keep intermediate data in on-chip GPU memory. While originally optimized for GPUs, recent work has explored FlashAttention on CPUs with vector units to improve transformer inference~\cite{titopoulos2025vectorizedflashattentionlowcostexponential, llama_cpp}. 

Modern vector processors with vector-length-agnostic Instruction Set Architectures (ISAs), such as the RISC-V Vector Extension (RVV) and the ARM Scalable Vector Extension (Arm SVE), support variable vector lengths. This makes them attractive for HPC and AI workloads, which can exploit the SIMD parallelism exposed by these architectures. However, FlashAttention exposes limited vector-level parallelism along the head dimension. 
Queries, Keys, and Values have dimensions N$\times$D, where $N$ is the sequence length, and $D$ is the head dimension. The FlashAttention output has the same shape, $N\times D$, with the size of $N$ varying widely (collapsing to 1 during token generation) and $D$ being small, typically 64 or 128 (see Table \ref{tab:NetworkModels}).
Existing implementations of FlashAttention on RISC-V, ARM, and x86 follow a restrictive design choice: the vector length (VL) is tied to $D$. The vectorized FlashAttention implementations in llama.cpp are all limited to $VL \leq D$~\cite{llama_cpp}. Titopoulos et al. ~\cite{titopoulos2025vectorizedflashattentionlowcostexponential} vectorize FlashAttention on RVV specifically for $VL \leq D$. MEATTEN~\cite{FuMeAttenICS2024}, a NEON-based fused-attention implementation, vectorizes over the head dimension $d_k$, thereby imposing the same restriction. For 16-bit precision, these implementations are able to utilize vector lengths only up to 1024–2048 bits, while more aggressively quantized models utilize even less. This limits the scalability potential of FlashAttention on long vector architectures, where vector lengths exceed the head dimension, leading to underutilization of Single Instruction, Multiple Data (SIMD) resources. 



In this work, we bridge this gap by optimizing FlashAttention for vector architectures across a wide range of vector lengths, enabling efficient utilization of both conventional and long vector registers. We propose \textit{\FlashV{}}, a redesigned FlashAttention algorithm 
that exploits parallelism across attention heads. By reordering loops, applying loop unrolling, and applying inter-head packing, \FlashV{} maps multiple heads into a single vector register, enabling the efficient utilization of vector lengths exceeding the head dimension $D$. Loop unrolling further exposes instruction-level parallelism, allowing \FlashV{} to efficiently utilize vector registers and scale across both conventional and long vector lengths in prefill and decode stages. Our design natively supports MHA and GQA, avoiding redundant memory accesses for shared key-value. 

We implement \FlashV{} in the ggml framework~\cite{ggml} within llama.cpp~\cite{llama_cpp}, a framework commonly used as a research platform~\cite{VSeekPoster25, LiuLlMininLlama25}. We use vector intrinsics of the RISC-V ISA~\cite{riscvisaintrinsic} and the Arm SVE ISA\cite{arm_sve2} to vectorize our kernels. We evaluate \FlashV{} on decoder-only Large Language Models (LLMs) that operate in two stages: prefill (processes input tokens and computes the KV cache) and decode (generates one token at a time). 
We use three GQA models: TinyLlama~\cite{TinyLlama}, Llama 3.2~\cite{Llama-3}, and Qwen2.5~\cite{qwen2025qwen25technicalreport}, as well as Pythia-410M~\cite{biderman2023pythiasuiteanalyzinglarge}, to evaluate MHA. 
We validate the functional correctness of \FlashV{} with Arm SVE using QEMU to demonstrate portability across vector ISAs.
We evaluate \FlashV{} with RVV on a Banana Pi BPI-F3 RISC-V board and use gem5~\cite{Gem5} for RVV to assess scalability up to 8192-bit vector lengths. To reflect real hardware behavior, where a fixed number of vector lanes processing longer vector lengths increases per-instruction execution latency, we extend gem5's MinorCPU with vector length-proportional SIMD functional unit latencies, informed by the lane-based microarchitecture of Vitruvius+~\cite{vitrus}. Finally, we extend our analysis beyond the attention mechanism to quantify bottlenecks in quantized linear projection and feed-forward layers, showing that vector packing and masked reduction overheads significantly limit scalability with longer VLs.

%
The contributions of this paper are as follows:
\begin{itemize}[noitemsep, topsep=8pt]
    \item We propose \FlashV{}, a FlashAttention algorithm for vector architectures that exploits parallelism across attention heads and introduces inter-head packing, enabling effective utilization of vector lengths beyond the head dimension ($VL > D$). Combined with loop reordering, inter-head unrolling, maximized vector register reuse, and native MHA/GQA support, \FlashV{} achieves 22$\times$–42$\times$ speedup over scalar (non-vectorized) FlashAttention in prefill and 8$\times$–11$\times$ in decode at 512-bit vector lengths using gem5 on RVV. Validation on Banana Pi BPI-F3 confirms these gains with 12$\times$-14$\times$ over scalar FlashAttention. 
    

    \item We evaluate the scalability of \FlashV{} using \texttt{gem5}, establishing upper-bound speedups of $2\times$-$3\times$ in prefill and $1.2\times$ in decode across vector lengths up to 8192 bits. Using a latency-aware per-lane model, we show that configurations with up to 64 vector lanes and 4096-bit vectors retain strong scalability, while highlighting the vector-length overheads that limit performance beyond 4096-bit vectors. These findings provide actionable insights for hardware and vector-length-aware FlashAttention optimization on long-vector architectures, which have not been quantified previously.

\item We characterize the structural incompatibility between Q8\_0 quantization and long-vector execution in linear projection and feed-forward modules. Through microbenchmark analysis on RISC-V, we identify packing and masked reduction as the dominant bottlenecks, consuming 60\% of execution cycles at 2048-bit VL. Validation on Arm SVE at 2048-bit VL confirms that these overheads are architecture-agnostic, stemming from the interleaved weight-scale layout of Q8\_0, which introduces VL-dependent costs that prevent arithmetic amortization. This highlights a fundamental limitation for long-vector execution in quantized models.

\end{itemize}

The rest of this paper is organized as follows. Section~\ref{sec:background} provides background on transformer models. Section~\ref{sec: related work} views related work. Section~\ref{sec:optimizations} describes the algorithmic transformations of FlashAttention and self-attention, and analyzes linear projection and feed-forward layers under Q8\_0 quantization. Section~\ref{sec:methodology} presents the experimental platforms and setup. Sections~\ref{sec:evaluation} and~\ref{sec:EvaluationLPFF} evaluate our \FlashV{} and analyze structural incompatibilities between quantized models and long-vector execution. We conclude the paper in Section~\ref{sec:conclusion}.


\section{Background}
\label{sec:background}

\input{Background}
\section{Related Work}
\label{sec: related work}
\input{Related_Work}
\section{Algorithmic Optimizations}
\label{sec:optimizations}
\input{Method}

\section{Methodology}
\label{sec:methodology}
\input{Models}

\section{Evaluation of \FlashV{}}
\label{sec:evaluation}
\input{Results}
\section{Structural Limits to Long-Vector Scaling in Quantized Layers}
\label{sec:EvaluationLPFF}
\input{quantizationAnalysis}

\section{Conclusion}
\label{sec:conclusion}

\input{Conclusion}

\bibliographystyle{ACM-Reference-Format}
\bibliography{sample-base}



\end{document}

%% file: abstract.tex
Inference with transformer models on CPUs is increasingly important, especially for Small Language Models (SLMs), where vector architectures are emerging as a promising execution substrate. The attention module is a major bottleneck due to high memory bandwidth requirements; FlashAttention mitigates this by fusing operations to improve data locality and reduce intermediate memory traffic.
In this paper, we present \FlashV{}, a blocked FlashAttention for scalable vector architectures that adapts efficiently from short to very long vectors by exploiting parallelism across attention heads, inter-head packing to enable efficient utilization of vector lengths beyond the head dimension, and improving vector register utilization and memory access locality. We integrate \FlashV{} into ggml within \texttt{llama.cpp} and evaluate it on TinyLlama, Llama 3.2, Qwen2.5, and Pythia-410M using \texttt{gem5} and a Banana Pi BPI-F3. On the Banana Pi BPI-F3, we confirm that loop reordering and loop unrolling across attention heads are effective optimization principles, scaling performance gains with larger models and most pronounced with short contexts and during decoding.  
Simulation-based analysis shows that \FlashV{} achieves 22$\times$-42$\times$ speedup over scalar FlashAttention at 512-bit VL in prefill, with an additional 2$\times$-2.5$\times$ gain scaling to 64 lanes and 4096-bit VL. 
During decode, \FlashV{} achieves 8$\times$-11$\times$ speedup using 512-bit vector lengths over scalar FlashAttention, with performance showing diminishing sensitivity to vector width and lane count due to single-token, memory-bound execution. We further identify structural bottlenecks in Q8\_0 quantized linear layers that limit arithmetic amortization under long-vector execution, consistent across RVV and Arm SVE, indicating that current quantization formats pose a fundamental challenge to long-vector scalability.



%% file: Background.tex
In this paper, we consider decoder-only models that use attention mechanisms such as Multi-Head Attention (MHA) or its efficient variant, Grouped-Query Attention (GQA), which reduces key–value cache overhead by sharing keys and values across query heads. 
These models are built upon the same underlying self-attention mechanism.
\textbf{\textit{Self-attention}} lets each token attend to all others~\cite{vaswani2017attention}. Input embeddings are projected into Queries (Q), Keys (K), and Values (V), and attention is computed via scaled dot product and softmax: $\text{Attention}(Q, K, V) = \operatorname{softmax}\!\left( \frac{Q K^{\top}}{\sqrt{D}} \right) V$ where $D$ is the head dimension. Multi-head attention applies this in parallel across multiple heads to capture diverse relationships.
\textbf{\textit{FlashAttention}} is an optimized implementation of self-attention that reduces memory usage and improves computational efficiency~\cite{dao2022flashattentionfastmemoryefficientexact}. Instead of materializing the full attention matrix in memory, it tiles the computation across blocks, maintaining running maxima and normalization factors for numerically stable on-the-fly softmax, avoiding intermediate memory overhead.

\texttt{llama.cpp}~\cite{llama_cpp} is an open-source CPU inference framework built on \texttt{ggml}~\cite{ggml}, a tensor library supporting quantized operations and vectorized instructions across AVX, Arm NEON, Arm SVE, and RVV. In \texttt{llama.cpp}, the attention module is implemented using both self-attention and FlashAttention.


%% file: Related_Work.tex
Several works have focused on optimizing attention mechanisms. We first discuss algorithmic improvements to self‑attention, followed by FlashAttention variants.   
Rabe and Staats~\cite{rabe2022selfattentiondoesneedon2} introduce a memory-efficient self-attention algorithm that avoids $O(n^2)$ memory usage. The FlashAttention algorithm ~\cite{dao2022flashattentionfastmemoryefficientexact} follows similar principles: it is an optimized implementation of the self-attention to reduce the memory bandwidth. It was further optimized by Dao et al. 
~\cite{dao2023flashattention2fasterattentionbetter}, who proposed FlashAttention-2. 

We next discuss CPU-specific implementations of self-attention and FlashAttention. Zhang et al.~\cite{ZhangTransformersOnCPUs2023} propose NIOT, a framework for efficient inference using kernel fusion and tiling, evaluated on BERT and ViT on Xeon 6226R with AVX-512. XNNPACK~\cite{xnnpack} is a hand-tuned library for ARM and x86 CPUs that provides a fused SDPA operator with thread-level parallelism. 
Fu et al.~\cite{FuMeAttenICS2024} propose MEATTEN, a memory-efficient attention scheme for ARM NEON-based multi-core CPUs. Martínez et al.~\cite{MartinezEncoderLayer2024} optimize self-attention on ARM and RISC-V CPUs for encoder-only models such as BERT. Rodrigo et al.~\cite{VSeekPoster25} optimize LLM inference in \texttt{llama.cpp} for the many-core RISC-V Sophon SG2042~\cite{brown2024performancecharacterisation64coresg2042}. Garcia et al.~\cite{MARQUESGARCIA2026108242} benchmark LLM inference on the same platform using PyTorch with OpenBLAS and BLIS backends. Liu et al.~\cite{LiuLlMininLlama25} optimize key kernels (vector dot product and layer normalization) in \texttt{llama.cpp} on the Banana Pi BPI-F3 RISC-V platform. On RISC-V vector architectures, Titopoulos et al.~\cite{titopoulos2025vectorizedflashattentionlowcostexponential} propose a FlashAttention variant restricted to $VL \leq D$, evaluated on Gemma-2 and Qwen models using gem5. \texttt{llama.cpp} introduced a GEMM-based vectorized attention alongside a reduction-based implementation~\cite{llama_cpp}, both limited to $VL \leq D$. OneDNN exposes SDPA through a graph-level fusion API~\cite{onednn_sdpa}, requiring intermediate storage of Query–Key dot products. It also supports GQA~\cite{onednn_gqa} through a graph-level pipeline combining reshape, MatMul, scaling, masking, and softmax into a single kernel for x86 CPUs and Xe GPUs. However, this remains a high-level fusion approach targeting server-class systems, and does not address kernel-level vector optimization for RISC-V edge processors.


The existing work on FlashAttention for RVV has been explored only in scenarios where the vector length is less than or equal to the head dimension, leaving the potential of long vector configurations unexploited. In this work, we redesign FlashAttention to exploit long vector lengths by exploiting parallelism across multiple attention heads, with native GQA support to eliminate redundant K/V loads.
\looseness=-1

%% file: Method.tex
In this section, we first present the algorithmic optimizations in \FlashV{}, followed by optimizations in the self-attention implementation and an analysis of the performance characteristics of the quantized model.

\subsection{\FlashV{} Optimizations}
For each attention head, the Query (Q), Key (K), and Value (V) tensors have shape (N $\times$ D), where $N$ is the sequence length and $D$ is the head dimension. In the decode stage, this reduces to (1 $\times$ D). The output matrix has the same dimensions, N$\times$D and 1$\times$D, per attention head for the prefill and decode stages, respectively. 
The \texttt{ggml} library includes an initial implementation for FlashAttention based on the streaming softmax formulation of~\cite{rabe2022selfattentiondoesneedon2}. A high-level pseudo-code of the ggml vectorized (\texttt{ggml-vec}) implementation of FlashAttention is presented in Algorithm~\ref{NaiveFlash}. The kernels $kq\_vec\_dot$, $ggml\_vec\_scale$, and $ggml\_vec\_mad$ (lines 6, 17,26, and 35) vectorize over the head dimension (D) and are limited by its size. 
Since these loops operate over $D$, the maximum exploitable vector length is bounded by D elements. For the common case of $D=64$ or $128$ with \texttt{FP16}, this corresponds to 1024-bit or 2048-bit VLs. Any vector length beyond these thresholds introduces no additional parallelism within a single head, as there are insufficient elements to fill those vector registers. Therefore, this FlashAttention implementation saturates vector utilization once the vector length exceeds $D$, as additional vector width cannot be effectively exploited within a single attention head. Our preliminary experimental evaluation validates this observation. To address this limitation, we redesign the ggml FlashAttention implementation and introduce \FlashV{}, which exploits parallelism across attention heads to utilize longer vector registers. For scalable vector architectures, \FlashV{} supports two execution regimes based on the hardware vector length relative to the head dimension: $VL > D$ (enabling inter-head packing) and $VL \le D$ (no inter-head packing). Across these regimes, \FlashV{} combines inter-head packing (for $VL > D$) with cache- and register-aware optimizations, blocking, loop reordering, and loop unrolling over attention heads, and GQA-aware data reuse, to improve data locality, vector utilization, and instruction-level parallelism.



\begin{algorithm}
\small
\caption{\texttt{ggml-vec} FlashAttention in llama.cpp} \label{alg:vector_attention}
\label{NaiveFlash}
\begin{algorithmic}[1]

\State Loop {$j_1 \gets 0$ \textbf{to} $N*attention\_heads$ \textbf{step} 1} //N - sequence length
         \State check for causal masking
         \State $M=-INFINITY$ //Keep global maximum
         \State $Sum = 0.0$ //keep global sum
             \State Loop  {$ic \gets 0$ \textbf{to} $N$ \textbf{step} 1}
                      \For{$i \gets 0$, \textbf{to} $D$} //$ kq\_vec\_dot(D, \&s, K_{ptr}, Q_{ptr})$ 
                        \State $avl \gets setvl(D-i)$
                        \State $V\_Q = Q_{ptr}[0:avl]$ $V\_K = K_{ptr}[0:avl]$
                        \State $s = vfmv(vfredusum(fmul(V\_Q, V\_K)))$ 
                        \EndFor
                       \State $S \gets s * scale$
                       \State $M_{old} \gets M$ //Keep the old maximum in $M_{old}$
                       \State $ms \gets 1.0$ $vs \gets 1.0$
                        \If {$S>M$}    //check for new maximum 
                        \State $M = S$
                        \State $ms = expf(M_{old} - M)$  //Calculate rescaling factor
                        \For{$i \gets 0$, \textbf{to} $D$} //$//ggml\_vec\_scale(D, VKQ, ms)$
                        \State $avl \gets setvl(D-i)$
                        \State $VKQ = VKQ\_array[0:avl]$
                        \State $VKQ = fmul(VKQ, ms, avl)$  //VKQ rescale
                        \State $VKQ\_array[0:avl] =VKQ$
                        \EndFor                        
                        \Else
                        \State $vs = expf(S - M)$  //Compute softmax weight
                        \EndIf
                         \For{$i \gets 0$, \textbf{to} $D$} //$ggml\_vec\_mad(D, VKQ, V_{ptr}, vs)$
                        \State $avl \gets setvl(D-i)$
                        \State $VKQ = VKQ\_array[0:avl]$ $V\_V = V_{ptr}[0:avl]$
                        \State $VKQ = vfmadd(VKQ, V\_V, vs, avl)$  //accumulate VKQ
                        \State $VKQ\_array[0:avl] =VKQ$
                        \EndFor
                        \State $Sum = Sum*ms+vs$  //Online softmax sum
                        \State End Loop $ic$
                        \State $Sum_{inv} = 1/Sum$
                        \State $ggml\_vec\_scale(D, VKQ, Sum_{inv})$//normalize VKQ
                        \State $dst = VKQ$ 
                        \Comment{put result back}
                        
\end{algorithmic}
\end{algorithm}

\begin{figure*}[t]
  \centering
  \includegraphics[width=\textwidth]{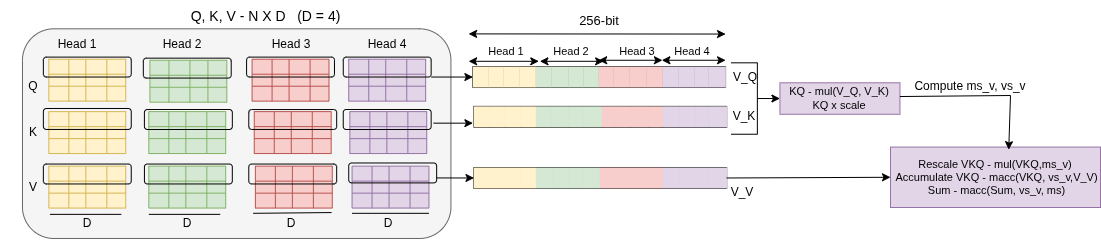}

  \caption{
     Inter-head packing into a single vector register to improve vector utilization across attention heads with FP16}
  \label{fig:MyApproach}
  
\end{figure*}

\begin{algorithm}
\small
\caption{ \FlashV{}  for VL > D}
\label{FlashAttention-I}
\begin{algorithmic}[1]
\State Initialize: $VL \gets \text{vsetmaxvl}()$, $avl \gets D$ $itr \gets VL/D$
\State Loop $j_1$ and $k_1$ to $N$ steps $block\_N$ and $block\_k$ //N - sequence length
        \State Loop {$j \gets 0$ \textbf{to} $block\_N$}
            \For{$i_1 \gets 0$ \textbf{to} $attention\_heads$ \textbf{step} $U$} //$U$ (unroll factor)
                \For{$i \gets 0$ \textbf{to} $R$} //R(register\_utilization\_factor) blocks of $itr$ 
                    \State $V\_Q[i] \gets Q_{ptr}[i_1, i*itr: VL]$
                \EndFor
                \If {$k_1 = 0$}
                \State $Sum_{v}[0{:}R]=0$, $VM[0{:}R]=-\infty$, $VKQ[0{:}R]=0$
                \Else   \State Load from previous $S_{array}$, $M_{array}$, $VKQ\_array$ 
                \EndIf
                
                \For{$ic \gets k_1$ \textbf{to} $k_1+block\_K$}
                    \State Check for Causal masking
                     \For{$i \gets 0$ \textbf{to} $R$}
                        \If{Grouped-Query Attention}
                            \State Load $K_{ptr}[ic, i*itr:avl]$ and Pack in $V\_K$ //avl=D
                        \Else
                             \State $V\_K[i] \gets vle (K_{ptr}[ic, i*itr:VL])$ //load full VL
                         \EndIf
                        \State $KQ[i] \gets vfmul(V\_Q[i], V\_K[i])$ //KQ\_vec\_dot
                        \State $KQ[i] \gets vfmul(KQ[i], scale)$ //Scale score
                        \State $VS[i] = vfredusum_m(mask,KQ[i])$ //scores
                        \State $VM_{old}[i] =VM[i]$ //keep old maximum
                        \State $VM = vmerge(VM, VS, mask)$ //new maximum
                        \State  $ms\_v[i]=exp\_v(VM_{old}[i], VM[i])$  //rescale factor
                        \State  $vs\_v[i]=exp\_v(VS[i],VM[i])$ //softmax weights
                        \State Load $V\_V$ and accumulate similar to $K$
                   
                        \Statex \hspace{\algorithmicindent} \hspace{\algorithmicindent}  //VKQ rescale 
                        \State $VKQ[i] \gets vfmul\_m(mask, VKQ[i], ms\_v[i])$
                        \Statex \hspace{\algorithmicindent} \hspace{\algorithmicindent}  //accumulate VKQ 
                        \State $VKQ[i] \gets vfmacc(VKQ[i], V\_V[i], vs\_v[i])$ 
                        \Statex \hspace{\algorithmicindent} \hspace{\algorithmicindent}  //Online softmax sum
                        \State $Sum_v[i] \gets vfmacc(vs\_v[i], Sum_{v}[i], ms\_v[i])$ 
                       
                    \EndFor
                \EndFor

                \State Update $S_{array}$, $M_{array}, VKQ\_array$
                \State Normalize $VKQ$ and store in $dst$
            \EndFor
\end{algorithmic}  
\end{algorithm}

\subsubsection{\FlashV{} optimizations when $VL > D$}
Given that vector parallelism over $D$ saturates at $VL = D$, individual attention heads cannot fully utilize vector registers when $VL > D$. We therefore exploit parallelism across attention heads to fill longer vector lengths. In this regime, \FlashV{} additionally introduces inter-head packing, mapping multiple attention heads into a single vector register to improve vector utilization under wide hardware configurations. We further improve efficiency by retaining intermediate results in vector registers across iterations, reducing memory traffic and improving data locality.
We implement blocked FlashAttention based on the non-vectorized \texttt{ggml} baseline, with blocking strategies inspired by FlashAttention-2~\cite{dao2023flashattention2fasterattentionbetter}, using cache-aware block sizes. A key observation is that attention heads are independent, as each head computes its own query ($Q$), key ($K$), and value ($V$) projections without cross-head dependencies. We exploit this independence as an additional source of parallelism by processing multiple heads simultaneously.
Analysis of the \texttt{ggml} memory layout shows that $Q$, $K$, and $V$ tensors across heads are stored contiguously. However, the baseline loop order does not exploit this structure. By reordering loops over attention heads, \FlashV{} improves spatial locality and enables more unit-stride-friendly memory accesses, allowing multiple heads to be mapped efficiently into a single vector register. As illustrated in Figure~\ref{fig:MyApproach}, four FP16 elements across four heads can be packed into a single vector register(256-bits), fully utilizing vector capacity.
To further improve utilization, we apply loop unrolling over attention heads, which increases vector register occupancy and exposes instruction-level parallelism (ILP) by keeping multiple vector registers active concurrently. Together, loop reordering and unrolling improve performance along two dimensions: (i) increasing effective vector width through inter-head packing, and (ii) improving instruction-level parallelism and memory latency hiding.


Algorithm~\ref{FlashAttention-I} shows the pseudo code for \FlashV{} for RVV when the vector length is greater than D. Line 1 initializes three key parameters: $VL$ is set to the maximum available vector length, $avl$ sets the vector length to $D$ (limit operations to $D$), and $itr=VL \div D$ determines how many attention heads fit within a vector register. The outer loops in line 2 tile the computation over $seq\_length$ in blocks of $block\_N$ and $block\_K$, selected to reduce memory traffic and improve data reuse. 
Loop over $attention\_heads$ is reordered to line 4 and incremented by a loop unroll factor $U$, to expose contiguous heads that enable inter-head packing and improve vector register utilization.
Lines 5–7 load Q into vector registers to maximize register reuse within the innermost loop and reduce memory bandwidth pressure. Lines 8-12 are used to keep $S_{array}$, $M_{array}$, and $VKQ\_array$  the global sum, global max, and global VKQ to support the online softmax operation~\cite{rabe2022selfattentiondoesneedon2}. They loaded in vector registers named as $sum_v$, $VM$, and $VKQ$.

In the inner-most loop, Keys are packed in lines 16-20 based on whether the model uses GQA or MHA. 
In the MHA case, K and V are contiguously aligned in memory and loaded using the $vle$ vector instruction in line 19. For grouped-query attention or multi-query attention, K and V values based on the $avl$ are loaded and replicated in vector registers using the \textit{vslideup} vector instruction. If the attention heads to KV head ratio is 4:1, i.e., four attention heads are shared among KV heads, then only a single load instruction is required to load K, after which the \textit{vslideup} instruction is used to replicate it across the full vector register. A similar packing mechanism is applied when loading $V$. 
Lines 21–31 implement core attention computation in a fully vectorized manner. The dot product between $V\_Q$ and $V\_K$ is computed and scaled in lines 21-22. The running maximum is updated in lines 23-25, from which the rescaling factor $ms\_v$ and the softmax weight $vs\_v$ are derived using the vectorized exponential function $exp\_v()$ in lines 26-27. We use $exp\_v$ from SIXTE~\cite{expf_vec}, modified to handle FP16 data precision, infinity values, and masking. The accumulated output $VKQ$ is rescaled and updated using \texttt{vfmul} and \texttt{vfmacc} vector instructions in lines 29-30, and the running softmax sum $sum_v$ is updated in line 31. After all tiles are processed, $VKQ$ is normalized and written to $dst$, completing the attention computation.
\paragraph{Tunable parameters in Algorithm~\ref{FlashAttention-I}}
The block sizes $block\_N$ and $block\_K$ are tuned to better fit within the L2 cache. In our implementation, $U$ is decided based on the vector length, the precision in which elements are stored, the head dimension, and the number of attention heads. Since the number of $attention\_heads$ varies across models, we must also consider it when deciding on the loop-unrolling. We set the unrolling factor $U$ as $U = \frac{VL}{D \cdot \text{b}} \times R $. Here, $b$ is the element bit-width and $R$ is the register utilization factor, which captures the additional loop unrolling enabled by available vector registers and the number of $attention\_heads$. For example, in TinyLlama and Llama 3.2, the head dimension is 64, so using $attention\_heads = 8$ maps cleanly onto an 8192-bit vector width (8 $\times$ 64 $\times$ 16 bits). In these models, the total number of heads is 32; therefore, after processing the first 8 heads, the remaining 24 heads allow increasing unrolling from 8 to 32 while still maintaining full utilization of the 8192-bit vectors, by setting $R$ to 4. 

\subsubsection{\FlashV{} optimizations when $VL \leq D$}
In this regime, vector parallelism operates within individual attention heads. Unlike the $VL > D$ regime: no inter-head packing is applied, $Q$, $K$, and $V$ are loaded according to the available vector length, and intermediate $VKQ$ results are accumulated and written back to memory between iterations, consistent with \texttt{ggml-vec} (Algorithm~\ref{NaiveFlash}). \FlashV{} extends this with optimizations shared across both regimes: (i) cache-aware blocking tuned to the cache hierarchy, consistent with Algorithm~\ref{FlashAttention-I}; (ii) loop reordering over attention heads; (iii) loop unrolling by factor $U$ on attention heads to enhance parallelism and maximize vector register utilization; and (iv) GQA support, where keys and values are loaded once and reused across shared attention heads to reduce memory bandwidth pressure.  

\subsection{Self-attention Optimizations}
For the self-attention implementation, ggml provides a blocked implementation involved in llama.cpp to accelerate the transformers on CPUs. Drawing on the optimization principles of \FlashV{}, we demonstrate that loop reordering and inter-head parallelism are broadly applicable to self-attention kernels, improving vector utilization across the QK$^{\mathsf{T}}$, softmax, and QKV accumulation stages during prefill computation.

 In the case of $QK^{\mathsf{T}}$, we apply vectorization across the head dimension to exploit data-level parallelism and enable contiguous memory accesses. Tokens in a Query are stored contiguously, allowing multiple tokens from a single head to be packed into long vector registers. For the Keys, one row of $D$ elements is loaded from memory and packed into long vector registers using the \texttt{vslideup} intrinsic, since computation is performed for one head at a time if $D<VL$. Multiple \texttt{vfmul\_vv} instructions then multiply the packed Query vectors with the loaded Key row, enabling maximum reuse of the Queries. For example, at 8192-bit VL, 32 FP16 Query rows of dimension $D=64$(as in TinyLlama) are packed across 4 vector registers per attention head, each multiplied against a single Key row, maximizing Key reuse within the innermost loop, amortizing the memory load cost of K across multiple token computations.

 In softmax, masking is applied to the attention score matrix, followed by the softmax operation. The score matrix has dimensions $N \times N$, where $N$ is the sequence length, allowing efficient use of long vector lengths. We reorder the attention-head loop and apply a loop unrolling factor of 16, balancing register usage and avoiding spilling. Since tokens across heads share the same mask, loaded data can be reused effectively. For $QK^{\mathsf{T}}V$, dimensions for the calculated attention weight matrix and Value matrix are  N$\times$N and N$\times$D, respectively, for the prefill stage. Here, N is large enough to utilize the long VLs. We apply a loop unrolling factor of 16, chosen to balance register utilization and avoid spilling, over the score matrix rows, multiplying 16 rows of the score matrix against a single Value row to produce 16 output results simultaneously, maximizing Value reuse within the innermost loop. Increasing loop unrolling from 16 to 32 degrades performance due to register spilling.

\subsection{Linear Projection and Feed-Forward }
Both prefill and decode consist of an input embedding, multiple decoder blocks, an output embedding, and a sampling layer. Each block includes Q/K/V projections, attention, and a feed-forward module. We evaluate TinyLlama on a Banana Pi BPI-F3 using a non-vectorized llama.cpp compiled with \texttt{LLVM} compiler. Llama-bench profiling shows FlashAttention dominates prefill (50\% per layer) but contributes only (10\%) in decode, where linear projections and feed-forward layers dominate. Therefore, in addition to \FlashV{}, we evaluate these modules for a complete decode-stage analysis.

Both modules are linear projections(input multiplied by a learned weight matrix). In decode, single-token inputs reduce them to General Matrix-Vector (GEMV)  operations implemented via \texttt{vec\_dot} in llama.cpp. We evaluate 8-bit GGUF Q8\_0 quantization, where weights are stored in 32-element blocks with symmetric INT8 values and a per-block FP16 scale (no zero-point). This layout follows an Array of Structures (AoS) format, with 32 INT8 weights followed by a 2-byte scale. On RISC-V, this AoS layout limits long-vector execution because contiguous loads interleave weights and scales, reducing effective vector utilization. To fully utilize wider vector configurations, the scattered 8-bit weights must be reorganized into contiguous vector registers (packing), and partial results must be reduced at the block level (unpacking). We explore mechanisms for efficient packing and reduction to enable execution under longer VLs and we discuss the challenges under Section~\ref{sec:EvaluationLPFF}.

%% file: Models.tex
\subsection{Experimental Platform} 
We use gem5~\cite{Gem5} 24.0.0.1, a cycle-accurate simulator configured with the RISC-V in-order \textit{RiscvMinorCPU} model with RVV v1.0 support. 
The simulated memory subsystem uses DDR3-1600 with a per-core bandwidth of 12.8 GB/s, within the same order of magnitude as the Intel Xeon Max 9480 with HBM (~19 GB/s)~\cite{mccalpin2023bandwidth}, ensuring that the results reflect realistic bandwidth constraints.
The CPU model includes two levels of cache, configured similarly to AMD Zen~4 processors~\cite{AmdZen4}. The main parameters are summarized in Table~\ref{tab:system_config}. 

 \begin{table}[t]
\centering
\caption{Simulated system configuration}
\label{tab:system_config}
\begin{tabular}{|c|c|}
\hline
\textbf{Component} & \textbf{Specification} \\
\hline
Core frequency     &  2GHz\\\hline
Vector length      & 512-bits to 8192-bits\\\hline
Memory type        & DDR3-1600 \\\hline
Memory bandwidth   & 12.8 GB/s per core \\\hline
Cache hierarchy    & L1: 64 kB; L2: 1 MB \\
\hline
\end{tabular}
\vspace{-1em}
\end{table}
 We note that in gem5’s MinorCPU model, the MinorDefaultFloatSimdFU (a floating-point and SIMD functional unit) executes all operations with a fixed latency of 6 cycles, independent of operation type or vector width. Since the out-of-order \textit{RiscvO3CPU} model could not complete simulations due to memory exhaustion, we instead scale functional unit latencies in the \textit{RiscvMinorCPU} model to better reflect real hardware behavior. 
 
We scale SIMD FU latencies in the \textit{RiscvMinorCPU} using $opLat=startup+\max\left(0,\left\lceil \frac{VL}{PhysicalWidth} \right\rceil - 1\right)$, where $PhysicalWidth$ represents the physical throughput per cycle of the vector unit to model realistic vector latency.  We use 512 bits/cycle as the reference throughput, modeled as the aggregate throughput of Vitruvius+'s 8 lanes operating at 64 bits per lane per cycle~\cite{vitrus}. We derive the base latencies from \texttt{O3\_ARM\_v7a.py} in the gem5 configurations, as summarized in Table~\ref{tab:latency_comparison}. The model extends to wider designs: Ara2 with 16 lanes gives 1024 bits, and AraXL with up to 64 lanes gives 4096 bits~\cite{ara2Perotti_2024, araxlPurayil_2025}. The startup term accounts for the FU pipeline depth at the base vector width; additional passes beyond the first each contribute one extra cycle.

To validate the correctness, we use QEMU 9.0.4 configured with \texttt{rv64,v=true,zvfh=true,zfhmin=true,vext\_spec=v1.0}. We note that gem5 25.0.0.1 produced inaccurate results despite identical cycle counts for RVV; we therefore use gem5 24.0.0.1 throughout.\footnote{Cycle counts remain identical between versions.} Finally, we evaluate \FlashV{} on the Banana Pi BPI-F3, which supports RVV v1.0 with a 256-bit vector length. For the Arm SVE packing analysis in linear projection and feed-forward layers, we use gem5 25.0.0.1 configured with the same configuration as the RVV setup. 
\begin{table}[!htbp]
\centering
\small
\caption{Comparison of MinorCPU fixed latencies, and ARM OoO base latencies (O3\_ARM\_v7a.py), with VL-scaling indicated for each operation class.}
\begin{tabular}{|c|c| c| c|}
\hline
\textbf{Op Class} & \textbf{MinorCPU } & \textbf{O3\_ARM  } & \textbf{VL-Scaled} \\
& Fixed (cycles) & base (cycles) &  \\
\hline
SimdFloatAdd        & 6 & 5 & Yes \\
\hline
SimdFloatAlu        & 6 & 5 & Yes \\
\hline
SimdFloatMult       & 6 & 3 & Yes \\
\hline
SimdFloatMultAcc    & 6 & 5 & Yes \\
\hline
SimdFloatReduceAdd  & 6 & 4 & Yes \\
\hline
FloatAdd            & 6 & 5 & No  \\
\hline
FloatCvt            & 6 & 5 & No  \\
\hline
FloatMult           & 6 & 4 & No  \\
\hline
FloatMultAcc        & 6 & 5 & No  \\
\hline
\end{tabular}

\label{tab:latency_comparison}
\vspace{-1em}
\end{table}

\subsection{Experimental Setup}
In this work, we evaluate \FlashV{} in both the prefill and decode stages across four models: three Grouped-Query Attention (GQA) models, TinyLlama~\cite{TinyLlama}, Llama 3.2~\cite{Llama-3}, and Qwen2.5~\cite{qwen2025qwen25technicalreport}, and one Multi-Head Attention (MHA) model, Pythia-410M~\cite{biderman2023pythiasuiteanalyzinglarge}. Their dimensions are summarized in Table~\ref{tab:NetworkModels}. Since RISC-V lacks BF16 support, our implementation uses FP32 and FP16 precision, consistent with recent studies~\cite{FuMeAttenICS2024, MartinezEncoderLayer2024}. We apply Q4\_K\_M and Q8\_0 (4-bit and 8-bit block quantization schemes, respectively) quantization in linear projection and feed-forward layers; by the time data reaches FlashAttention, tensors are already expanded to native FP16 or FP32 formats. All experiments use a batch size of 1, corresponding to a single inference request.
 
 We use vector intrinsics from the RISC-V ISA~\cite{riscvisaintrinsic} to vectorize the attention implementation in llama.cpp for RVV. We use the EPI fork of the LLVM Clang cross-compiler version 21.0.0 for RVV~\cite{llvm-epi}, with \texttt{-O3} and \texttt{-rv64gcv\_zvfh} compiler flags to enable the support of half-precision with vector intrinsics to compile llama.cpp with our \FlashV{}. We use GCC cross-compiler version 15.2.1 for Arm SVE, with \texttt{-O3} and \texttt{-march=armv8.2-a+sve2} compiler flags. Functional correctness of our \FlashV{} on Arm SVE is verified using qemu-aarch64 version 8.2.2; performance evaluation is conducted on RVV only.
 We integrate our implementation with ggml in llama.cpp project\footnote{\url{https://github.com/ggml-org/llama.cpp}, commit \texttt{5fc7cb2021231998f7fbeefbdf21feddf1b4d746}}  
All models are evaluated using the \texttt{llama-bench} benchmarking tool from llama.cpp. 

\paragraph{\textbf{Baseline}:} We evaluate \FlashV{} against three baselines. Here, \texttt{ggml-scalar} is the non-vectorized FlashAttention implementation provided by ggml in llama.cpp. \texttt{ggml-vec-fp16} is the vectorized FlashAttention implementation provided by ggml in llama.cpp. \texttt{ggml-vec-fp32} is derived from \texttt{ggml-vec-fp16} by explicitly setting the FP32 execution path for all vectorized functions. Titopoulos et al.~\cite{titopoulos2025vectorizedflashattentionlowcostexponential}  also provide a FlashAttention implementation for RVV; however, their tensor layout is incompatible with llama.cpp, precluding a direct comparison.  

\begin{table}
    \centering
    \caption{Dimensions of Decoder-only LLMs Used in Our Evaluation}
    \begin{tabular}{|c|c|c|c|c|c|}
    \hline
        Model  & Attention  & KV  & Hidden &Head & Size\\
    
         Variants &   Heads &  Heads & Size &Dimension & \\
         & & & & (D) &\\
          \hline
        TinyLlama & 32 & 4 &2048 &64 & 1.1B\\
         \hline
        Llama 3.2 & 32 & 8 &2048 & 64 & 1B\\
         \hline
        Qwen2.5 & 16 & 2 & 2048 & 128 & 3B\\
         \hline
         Pythia & 16 & 16 & 1024& 64 & 410M\\
         \hline
    \end{tabular}
    
    \label{tab:NetworkModels}
    \vspace{-1em}
\end{table}



%% file: Results.tex
In this section, we present the impact of our optimized FlashAttention on the Banana Pi BPI-F3. We evaluate both the performance of the attention module and end‑to‑end inference using TinyLlama, Llama‑3.2, Qwen2.5, and Pythia‑410M. We further use gem5@RVV to analyze \FlashV{} with a 512‑bit and larger VLs to assess its single‑core scalability. In gem5, we simulate a single layer from both the prefill and decode stages. Since all layers operate on the same tensor dimensions, this is sufficient to capture behavior across the network while significantly reducing simulation time, especially in the prefill stage. When reporting FlashAttention performance, we report results for the attention module only; preprocessing and feed‑forward modules are excluded from these measurements. All baselines are evaluated against llama.cpp   \footnote{\url{https://github.com/ggml-org/llama.cpp}, commit \texttt{5fc7cb2021231998f7fbeefbdf21feddf1b4d746}}.

\subsection{\FlashV{} on Banana Pi BPI-F3}

We optimize \FlashV{} with FP32 and FP16 implementations on RVV, which can efficiently utilize both shorter and longer vector lengths. On the Banana Pi platform, the vector length is 256 bits (i.e., $VL \leq D$), eliminating the need for packing across heads. However, we unroll the attention heads by a factor of loop unroll $U$ as discussed in section~\ref{sec:optimizations}.  We unroll attention heads by a factor $U$, tuned over $\{1, 2, 4, 8\}$; performance degrades for $U > 4$, so we use $U=4$ throughout. We use the TinyLlama 1.1B, Llama 3.2,  Qwen2.5 models, and Pythia-410M for our analysis. 

  \begin{figure}[t]
  \centering
  \includegraphics[width=0.9\linewidth]{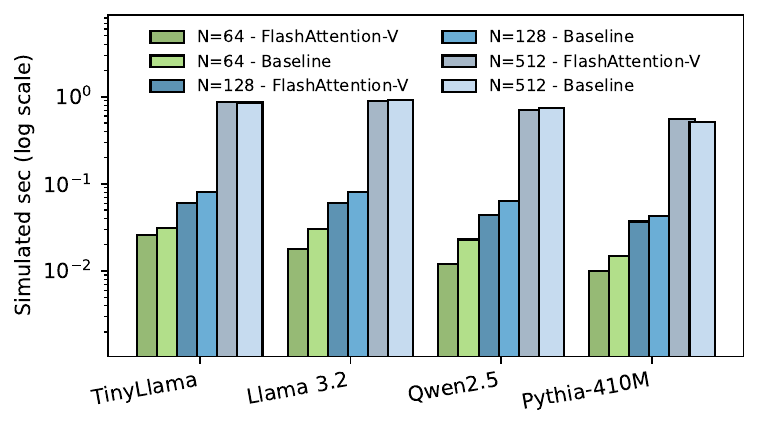}
  \caption{Prefill Performance Comparison of \FlashV{} and \texttt{ggml-vec-fp16} }
  \label{fig:model_results}
  \vspace{-1em}
\end{figure}

We first compare our \FlashV{} (FP32) in the \textit{prefill stage}. We compare against \texttt{ggml-scalar} and \texttt{ggml-vec-fp32} in the prefill stage. Our analysis shows that \FlashV{} achieves 12$\times$ and 14$\times$ speedup over \texttt{ggml-scalar} for TinyLlama and Qwen2.5, respectively, and a 3.7$\times$ speedup over \texttt{ggml-vec-fp32} for both models. Compared to an optimized self-attention, \FlashV{} achieves a $1.4\times$ speedup for TinyLlama 1.1B, confirming the benefit of the FlashAttention algorithm.

Further, our analysis in Figure~\ref{fig:model_results} shows that in the prefill stage, \FlashV{} with FP16 matches \texttt{ggml-vec-fp16} at $N=512$ and outperforms it at shorter sequence lengths $(N\le128)$. For $N=512$, \FlashV{} performs comparably to \texttt{ggml-vec-fp16} on TinyLlama and Llama 3.2; on Qwen2.5, it achieves approximately 5\% higher throughput, while on Pythia‑410M it yields about 7\% lower throughput at the same sequence length. 
This regression in Pythia-410M arises because it has fewer attention heads (16 vs 32) and uses MHA without key-value sharing, while Qwen2.5 benefits from a larger head dimension (128 vs 64), which provides more elements per head to amortize blocked computation and improve vector utilization.

We additionally evaluate different sequence lengths, namely $N \in \{64, 128, 256\}$, and \FlashV{} achieves an average speedup of 1.2$\times$–1.5$\times$ over \texttt{ggml-vec-fp16}, with the largest gains at $N=64$ and $N=128$ for TinyLlama, Llama 3.2, and Pythia‑410M. For Qwen2.5, \FlashV{} delivers 1.5$\times$–2$\times$ speedup over \texttt{ggml-vec-fp16} at $N=64$ and $N=128$. Overall, \FlashV{} is most effective at short sequence lengths $(N\le128)$ 
conditions under which our blocked design maximizes data reuse and reduces memory pressure. These small token sizes align naturally with practical edge deployment scenarios; our experiments are conducted on the Banana Pi BPI-F3, a representative emerging IoT and embedded inference platform. On such resource‑constrained devices, prefill sequence lengths of $\leq$128 tokens are common in practice and consistent with prior edge inference studies that adopt $N\leq128$ as a baseline~\cite{wangriscvextensionforsoftmax, TianLLminferenceedge2025usenix, aryaLLMonEdge2025}.

In the \textit{decode stage} with both FP16 and FP32, our implementation achieves  4$\times$ - 5$\times$ speedups over \texttt{ggml-scalar}  for the TinyLlama, Llama 3.2,  Qwen2.5, and Pythia-410M models, respectively. Furthermore, \FlashV{} with FP16 and FP32 delivers $\sim$2$\times$ speedup over \texttt{ggml-vec-fp16} and \texttt{ggml-vec-fp32} for all models. Ultimately, this confirms that \FlashV{} significantly accelerates both the prefill and decode stages, proving highly effective across all short-sequence regimes. We note that all optimizations discussed for \FlashV{} are transferable to other vector architectures.

We next evaluate the \textit{end‑to‑end performance} of the prefill and decode stages. For this, we integrate \FlashV{} into \texttt{llama.cpp}~\footnote{\url{https://github.com/ggml-org/llama.cpp}, 
commit \texttt{d0061be838809230db7a4edf62bc9a098025ba98}}. In this version, \texttt{llama-simple} serves as a minimal inference application with FlashAttention enabled by default. We verify the correctness of \FlashV{} by confirming that the generated tokens match those of the ggml baseline implementation.
Using this setup, \texttt{llama-simple} achieves end-to-end throughput of 6~T/s and approximately 3~T/s for TinyLlama\_Q4\_K\_M and TinyLlama\_Q8\_0, respectively, across both prefill and decode stages. Using \texttt{llama-bench} for the prefill stage (128 tokens and 8 threads), \FlashV{} achieves a 4\% improvement over the ggml vectorized baseline for Qwen2.5\_Q4\_K\_M (3.52~T/s), while remaining comparable for TinyLlama\_Q4\_K\_M (8.5~T/s) and Llama-3.2\_Q4\_K\_M (6~T/s). We note that the \texttt{llama-bench} evaluation for Qwen2.5\_Q8\_0 did not complete under the tested configuration. For the decode stage (128 generated tokens, 8 threads), \FlashV{} achieves performance comparable to the baseline for Qwen2.5\_Q4\_K\_M (2~T/s), TinyLlama\_Q4\_K\_M (5.7~T/s), and Llama-3.2\_Q4\_K\_M (5.3~T/s). These results are consistent with the overall system behavior: linear projection and feed-forward layers are identical across implementations, and the impact of FlashAttention is dominant in the prefill stage (approximately 50\% of per-layer execution) but limited in decode (approximately 10\%), leading to modest end-to-end gains.

\subsection{\FlashV{} using gem5} The ggml vectorized FlashAttention is limited to VL $\leq$ D and does not scale beyond the head dimension. To utilize longer vector lengths, \FlashV{} employs parallelism across attention heads, using inter-head packing to map Q, K, V rows from multiple heads into a single vector register. We evaluate this using gem5 on RVV across vector lengths from 512 to 8192 bits.


We first tune the tunable parameters(loop unrolling factor and block size) of \FlashV{} with a single prefill layer of TinyLlama at a sequence length $N=512$ and a 512-bit vector length using gem5@RVV with the default in-order MinorCPU configuration. For algorithm $VL \le D$, we achieve maximum utilization of the vector registers with unrolling factor $U=4$. Increasing $U$ to $U=8$ leads to an 8\% performance degradation at a 512-bit vector length.
We then tune the block sizes $block\_N$ and $block\_K$. Starting from sizes of $64\times64$ as the baseline, $128\times128$ achieves near-identical performance ($0.97\times$), while $16\times16$ and $16\times128$ reduce performance to $0.50\times$ and $0.89\times$, respectively.
We therefore set $U=4$ and $block\_N\times block\_K = 64\times64$ for all subsequent experiments. 

We evaluate our \FlashV{} against \texttt{ggml-scalar}, \texttt{ggml\-vec-fp32}, and \texttt{ggml-vec-fp16} for a prefill and decode stage for TinyLlama. For the \textit{prefill stage}, \FlashV{} with FP32 achieves speedup of 24$\times$ and 4$\times$ compared to \texttt{ggml-scalar} and \texttt{ggml-vec\-fp32}, respectively. With FP16, \FlashV{} achieves a 28$\times$ speedup over \texttt{ggml-scalar}. For the \textit{decode stage}, where the model generates one token at a time (making $Q$, $K$, $V$ tensors of size $1 \times D$), achieving parallelism to fill the vector lengths is more challenging.  In the decode stage, \FlashV{} with FP32 achieves a 7.6$\times$ speedup over \texttt{ggml-scalar} and a 3.8$\times$ speedup over \texttt{ggml\-vec-fp32} on TinyLlama. We note that we were unable to simulate \texttt{ggml-vec-fp16} with 
gem5 due to unsupported instructions.\footnote{Specifically, gem5 does not support LMUL=8 for widening instructions when simulating \texttt{ggml-vec-fp16}.}

\begin{table}
    \centering
    \caption{Execution time for 1 FlashAttention using gem5@RVV with 512-bits VL for TinyLlama}
    \begin{tabular}{|c|c|c|}
    \hline
        Version & Prefill stage & Decode stage  \\
        \hline
        Non-vectorized  & 21.38   & 0.000529\\
        \hline
        Vectorized by ggml with FP32 & 3.67 & 0.000300 \\
        \hline
        \FlashV{} with FP32 & 0.89  & 0.000077\\
        \hline
        \FlashV{} with FP16 & 0.78  & 0.000070 \\
        \hline
    \end{tabular}
    
    \label{tab:flashatten}
     \vspace{-1em}
\end{table}

\paragraph{FP16 vs FP32}
We compare \FlashV{} with FP16 and FP32 at the smallest and largest VLs in our simulated environment. FP16 halves storage relative to FP32, allowing twice as many elements per vector register.
We observe that \FlashV{} with FP16 yields $\sim$2$\times$, over \FlashV{} with FP32 with 8192-bit VL, but only 11\% at 512-bit vectors. For a decode stage, we observe a $\sim$1.5$\times$ speedup with 8192-bit vectors, while only a 7\% improvement is seen with 512-bit vectors. For 512-bit vector lengths (VL<=D), algorithmic implementation uses the same unrolling factor in both FP32 and FP16, as no packing across multiple heads is required. FP16 enables performing arithmetic operations on twice the number of elements, but the number of attention scores computed remains unchanged. Consequently, the benefit of FP16 is limited by the softmax bottleneck rather than the arithmetic throughput.

\paragraph{Self-attention vs \FlashV{}}
We next evaluate the performance of \FlashV{} over self-attention, after optimizing self-attention to utilize longer vector lengths, with FP16. We validate that our optimization renders self-attention scalable to longer vectors, yielding a speedup of $1.6\times$ when scaling the vector length from 512 bits to 8192 bits. 
Our results show that \FlashV{} gives a speedup of 1.3$\times$ and $\sim$2.4$\times$ over the optimized self-attention algorithm with 512-bit and 8192-bit vector lengths, respectively, in the prefill stage, confirming the advantages of \FlashV{} across vector lengths. 

\subsection{Scaling \FlashV{}on different vector lengths}
 
Further, we analyse the scalability with different vector lengths using gem5@RVV. This enables us to assess the effectiveness of \FlashV{} across varying vector lengths. We note that we use \FlashV{} with FP16 for our scalability analysis with gem5. We first simulate \FlashV{} using the default operation latencies of the in-order CPU in gem5@RVV. We then evaluate performance using operation latencies scaled according to the number of vector lanes, as discussed in Section~\ref{sec:methodology}. For longer vector lengths ( $VL > D$), we use a register utilization factor  $R=4$ to calculate the unrolling factor $U = \frac{VL}{D \cdot \text{b}} \times R $. 

\begin{figure}[t]
  \centering
  \includegraphics[width=0.9\linewidth]{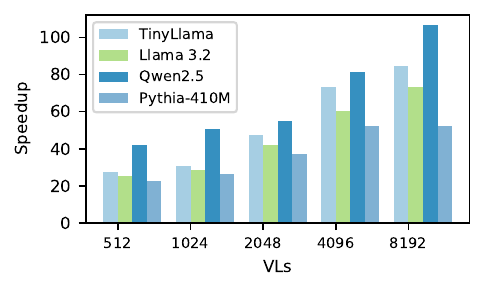}
  \caption{Prefill stage: Speedup across different vector lengths using gem5@RVV for TinyLlama, Llama 3.2, Qwen2.5, and Pythia-410M with 512 tokens, compared to \texttt{ggml-scalar}}
  \label{fig:Models-VLsonriscv}
  \vspace{-1em}
\end{figure}

\begin{figure*}[t]
  \centering
  \includegraphics[width=\textwidth]{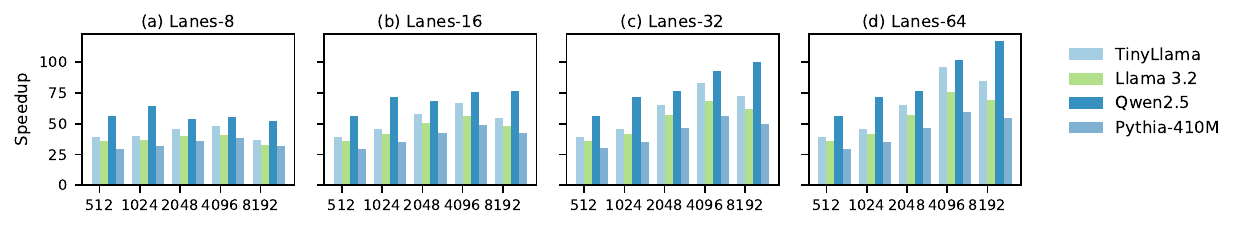}

  \caption{
     Prefill speedup across vector lengths (gem5@RVV, 512 tokens) for TinyLlama, Llama 3.2, Qwen2.5, and Pythia-410M, with VL-proportional operation latencies per vector lane, normalized to \texttt{ggml-scalar}.}
  \label{fig:Models-VLsonriscvLanes}
  \vspace{-1em}
\end{figure*}

\paragraph{Prefill stage}
Figure~\ref{fig:Models-VLsonriscv} shows that \FlashV{} consistently improves performance over \texttt{ggml-scalar} across 512-bit to 8192-bit vector lengths on TinyLlama, Llama~3.2, Qwen2.5, and Pythia-410M under a gem5 in-order model configured with constant operation latencies to establish an upper bound for achievable speedup. At the 512-bit vector length, \FlashV{} achieves speedups of 22$\times$–27$\times$ for TinyLlama, Llama~3.2, and Pythia-410M, and up to 42$\times$ for Qwen2.5. Increasing the vector length to 8192 bits yields an additional $\sim$3$\times$ speedup for TinyLlama and Llama~3.2, and $\sim$2.5$\times$ for Qwen2.5, demonstrating consistent benefits from wider vector utilization. The magnitude of these gains is governed by model architecture factors, such as per-head dimension and number of attention heads, which determine how effectively heads can be packed into vector registers. In particular, Qwen2.5 (head dimension 128) packs 4 heads per 8192-bit vector, whereas TinyLlama and Llama~3.2 (head dimension 64) pack 8 heads, resulting in better scalability. Pythia-410M, which is a smaller model, achieves its maximum speedup of 3$\times$ at 4096 bits, with full register utilization with $R=4$ and $U=16$. Beyond 4096 bits, the limited number of attention heads (16) does not allow the increase of the register utilization factor $R$, which is limited to $R=2$, resulting in the same unrolling factor of $U=16$.  Overall, our results confirm that \FlashV{} scales consistently with vector length across all models, with performance gains varying based on model architecture.

We next evaluate the scalability of \FlashV{} with operation latencies scaled according to the vector length, by parameterizing the vector unit across 8, 16, 32, and 64 lanes. Figure~\ref{fig:Models-VLsonriscvLanes} demonstrates that physical lane count is the primary determinant of performance scalability under realistic vector latency constraints. Lower lane counts exhibit early saturation: for the 8-lane configuration, performance degrades beyond 1024 bits, while 16- and 32-lane configurations extend this threshold to 4096 bits, with progressively improved scalability in intermediate ranges. In contrast, the 64-lane configuration achieves the best scalability, yielding 2$\times$-2.5$\times$  speedups when increasing the vector length from 512 to 4096 bits, for all models, with up to an additional 1.15$\times$ gain for Qwen2.5 on 8192 bits. Increasing the lane count from 8 to 64 widens the physical throughput from 512 to 4096 bits per cycle, accommodating longer vectors in fewer passes and significantly reducing the total \text{OpLat} penalty. Overall, our results indicate that 64 vector lanes paired with a 4096-bit VL achieve better scalability, successfully balancing theoretical throughput gains against VL-dependent latency overheads.\looseness=-1

Overall, under idealized fixed latencies, \FlashV{} scales consistently up to 8192-bit vector lengths. However, we demonstrate that under realistic latency scaling, where performance is governed by VL-dependent latency overhead of the vector functional unit, configurations with 64 vector lanes and 4096-bit vectors achieve the highest scalability among the studied designs, sustaining 2$\times$-2.5$\times$ speedup while avoiding the latency penalties that limit performance beyond 4096 bits. Our results also demonstrate that configurations with 64 vector lanes and 4096-bit vector lengths achieve comparable or higher speedups relative to the speedup achieved by constant-latency, indicating that \FlashV{} is not limited by operation latency (\text{OpLat}) in the regime of long vector widths when considering 64 vector lanes. \looseness=-1
\begin{figure}[t]
  \centering
  \includegraphics[width=\linewidth]{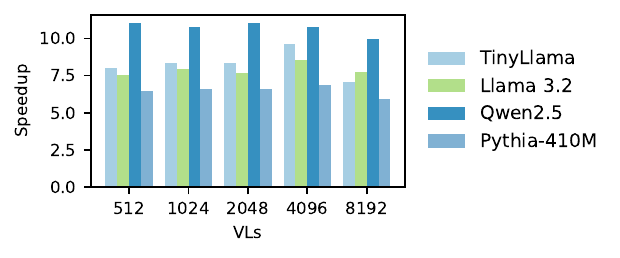}
  \caption{Decode stage: Speedup across vector lengths (VLs) on gem5@RVV for TinyLlama, Llama 3.2, Qwen 2.5, and Pythia-410M, relative to \texttt{ggml-scalar}}
  \label{fig:ModelsFlashonriscvGen}
  \vspace{-1em}
\end{figure}

\begin{figure*}[t]
  \centering
  \includegraphics[width=\textwidth]{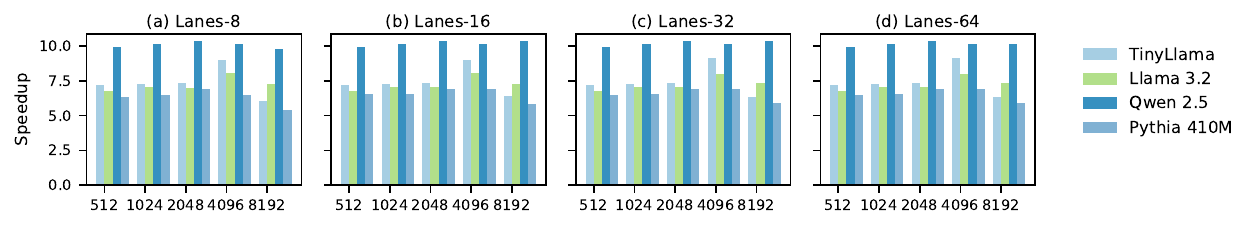}
  \caption{Decode stage: Speedup across different vector lengths (X-axis) using gem5@RVV for TinyLlama, Llama 3.2, Qwen2.5, and Pythia-410M, considering operation latencies per vector lane (baseline: \texttt{ggml-scalar})}
  \label{fig:ModelsFlashonriscvGenOpLat}
  \vspace{-1em}
\end{figure*}
\paragraph{Decode Stage} As discussed earlier, during the decode phase, only one token is generated at a time; thus, the $Q$, $K$, and $V$ vectors for the new token have a dimension of $1\times D$ per head. This inherent lack of parallelism makes it challenging to utilize longer vector lengths. Our \FlashV{} implementation bypasses this constraint by distributing computation across attention heads, aligning decode-stage execution with the only available source of parallelism in single-token generation.

Figure~\ref{fig:ModelsFlashonriscvGen} shows the performance of \FlashV{} across vector lengths from 512-bit to 8192-bit on TinyLlama, Llama~3.2, Qwen2.5, and Pythia-410M over \texttt{ggml-scalar}, under a gem5 in-order configuration with constant latency across vector lengths to represent the upper bound of achievable speedup. At 512-bit vector length, \FlashV{} achieves 8$\times$–11$\times$ speedup across all models. Scaling to 4096 bits yields a modest $\sim$1.2$\times$ improvement for TinyLlama and Llama~3.2, while Pythia-410M and Qwen2.5 show no significant additional gains. This saturation arises because the decode stage operates in a sequential single-token regime, which limits inter-token reuse and limits the amount of parallel work available per step.
When the vector registers become wider than a single attention head, the per-iteration workload is insufficient to amortize the overhead of packing/unpacking data into/from the wider vector registers, leading to diminishing returns from additional vector length scaling. \looseness=-1

Figure~\ref{fig:ModelsFlashonriscvGenOpLat} shows the speedup of \FlashV{} across increasing vector lengths under scalable operation latencies, compared to \texttt{ggml-scalar} during the decode stage. Under a fixed-latency model, small but observable differences appear across vector lengths, indicating minor sensitivity to vector scaling. However, with realistic, scalable operation latencies, these differences are significantly reduced, and performance becomes nearly invariant across vector lengths and lane counts. This behavior again owes to the fact that the decode stage processes only a single token per step, leaving the total vectorizable work too small for operation latency differences across configurations to affect execution time. Consequently, decode performance is effectively insensitive to \text{OpLat}, with only small variation across vector lengths and lane configurations. 

Overall, the optimizations applied in \FlashV{} benefit both the prefill and decode stages. In the decode stage, where the sequence length is fixed at 1, parallelism can only be extracted across attention heads. Because the computational workload per iteration is small, the overhead of data packing and managing wider vector registers is not fully amortized. Even under these constraints, \FlashV{} still achieves up to a 
1.2$\times$ speedup with 4096‑bit vector lengths for TinyLlama and Llama~3.2, demonstrating that its vector‑level optimizations remain effective even in the most parallelism‑constrained setting.


%% file: quantizationAnalysis.tex
Vectorizing the \texttt{vec\_dot} operation to exploit long vector lengths (VLs) in linear projection and feed-forward layers presents challenges distinct from attention kernels. In the ggml Q8\_0 format, each quantized block contains 32 INT8 weights prefixed by a single FP16 scale, stored contiguously. This interleaved layout disrupts vectorization: vector loads over weights also fetch scale values, requiring explicit separation prior to computation. We take ggml’s Q8\_0 vectorized implementation as a baseline and evaluate three strategies to leverage longer VLs.

First, we apply register packing using \texttt{vslideup}, concatenating multiple quantized blocks into a single vector register. This is followed by widening multiply–accumulate and a masked reduction (\texttt{vwredsum\_m}) to produce per-block results. To avoid iterative extraction via \texttt{vslidedown}, we rely entirely on masked reductions. Despite reducing arithmetic instructions, this approach degrades performance in TinyLlama by $\sim$1.5$\times$. To isolate the bottlenecks, we design a microbenchmark that replicates \texttt{vec\_dot} at projection dimensions and evaluates packing and reduction independently. On gem5 with a 2048-bit VL, packing via \texttt{vslideup} accounts for 28\% of cycles, masked reduction for 32\%, and vector MAC for 40\%. Thus, packing and reduction together consume 60\% of execution time, exceeding arithmetic cost and negating the benefits of longer VLs. 
We also evaluate packing on Arm SVE using \texttt{svcreate4}, \texttt{svst4}, and predicate-based \texttt{svld1}; among these, \texttt{svld1} performs best (as in llama.cpp for 512-bit VL), yet still incurs $\sim$ 20\% overhead with gem5 with a 2048-bit VL.

Second, we restructure the computation to vectorize across the output dimension \(M\), analogous to GEMV. This removes both packing and reduction overheads but introduces strided memory accesses (stride = 32) to respect quantization block boundaries. On the Banana Pi BPI-F3, this approach underperforms the baseline: non-unit strides degrade cache-line utilization, and increased memory latency outweighs the computational savings.

Third, we explore using LMUL=4 register groups with manual loop unrolling to map individual block loads into successive LMUL=1 segments, targeting up to 1024-bit effective VL. This avoids explicit packing entirely. However, the current \texttt{gem5} RISC-V vector model does not correctly simulate register-group type conversions when $\text{LMUL} \geq 4$, precluding reliable cycle-count validation. We identify this as a gap in current simulation infrastructure and leave hardware validation for future work.

We conclude that, under Q8\_0 quantization, linear projection and feed-forward modules do not benefit from long vector lengths in their current form. The limitation is structural: the interleaved scale–weight layout forces either non-unit-stride accesses or explicit packing, both of which introduce VL-dependent overheads that offset arithmetic gains. As quantized formats become the norm, the amortization argument for longer VLs weakens: when packing and reduction costs scale with VL alongside arithmetic, the overhead never gets amortized. Therefore, it becomes important to explore quantization formats or memory layouts that eliminate explicit packing overhead, would enable longer vector lengths to amortize arithmetic cost, and remain an open direction for future work.\looseness=-1

%% file: Conclusion.tex
This paper introduces \FlashV{}, a blocked FlashAttention algorithm for scalable vector architectures that exploits parallelism across attention heads to utilize longer vector lengths and maximize register utilization and reuse. We implement and evaluate \FlashV{} on RVV, and validate portability to Arm SVE via QEMU. Using gem5 on RVV at 512-bit VL, \FlashV{} achieves 22$\times$–42$\times$ speedup over non-vectorized FlashAttention in prefill and 8$\times$–11$\times$ in decode. Scalability analysis shows that 64 vector lanes paired with 4096-bit VL achieve optimal scalability, sustaining 2$\times$–2.5$\times$ speedup in prefill and 1.2$\times$ in decode, beyond which VL-dependent latency penalties outweigh throughput gains. The relative speedup of 8192-bit over 512-bit VL remains stable as sequence length grows from 512 to 4096 tokens, confirming consistent scaling behavior.
The core optimizations of \FlashV{}, loop reordering, and inter-head unrolling are transferable across execution strategies, as confirmed by a 5\% improvement when applied to the more recent tiled GEMM-based FlashAttention implementation~\footnote{\url{https://github.com/ggml-org/llama.cpp}, commit \texttt{5637536517ae4ed3eaa22b39c0d479e049097a9b}} in \texttt{llama.cpp} on the Banana Pi BPI-F3.
While \FlashV{} demonstrates strong attention kernel performance, linear projection and feed-forward modules under Q8\_0 quantization do not benefit from extended VLs due to structural layout constraints; exploring alternative quantization formats remains future work.